# Reading Comprehension Ability Test – A Turing Test for Reading Comprehension

Yuan Miao[1], Gongqi Lin[1], Yidan Hu[2], Chunyan Miao[2]

[1] Victoria University, PO Box 14428, Melbourne, VIC 8001, Australia
[2] Nanyang Technological University, Nanyang Ave, Singapore 639798

**Abstract**

Reading comprehension is an important ability of human intelligence. Literacy and numeracy are two most essential foundation for people to succeed at study, at work and in life. Reading comprehension ability is a core component of literacy. In most of the education systems, developing reading comprehension ability is compulsory in the curriculum from year one to year 12. It is an indispensable ability in the dissemination of knowledge. With the emerging artificial intelligence, computers start to be able to read and understand like people in some context. They can even read better than human beings for some tasks, but have little clue in other tasks. It will be very beneficial if we can identify the levels of machine comprehension ability, which will direct us on the further improvement. Turing test is a well-known test of the difference between computer intelligence and human intelligence. In order to be able to compare the difference between people reading and machines reading, we proposed a test called (reading) Comprehension Ability Test (CAT).CAT is similar to Turing test, passing of which means we cannot differentiate people from algorithms in term of their comprehension ability. CAT has multiple levels showing the different abilities in reading comprehension, from identifying basic facts, performing inference, to understanding the intent and sentiment.

1. **Reading Comprehension**

   *A Reading comprehension is an important human intelligence.*

   According to Wikipedia, reading comprehension "is the ability to process text, understand its meaning, and to integrate with what the reader already knows. Fundamental skills required in efficient reading comprehension are knowing meaning of words, ability to understand meaning of a word from discourse context, ability to follow organization of passage and to identify antecedents and references in it, ability to draw inferences from a passage about its contents, ability to identify the main thought of a passage, ability to answer questions answered in a passage, ability to recognize the literary devices or propositional structures used in a passage and determine its tone, to understand the situational mood (agents, objects, temporal and spatial reference points, casual and intentional inflections, etc.) conveyed for assertions, questioning, commanding, refraining etc. and finally ability to determine writer's purpose, intent and point of view, and draw inferences about the writer (discourse-semantics)."

   From the definition we can see that reading comprehension involves different types of abilities or intelligence, such as

- knowing the meaning of words,
- identifying the main thought of a passage,
- understanding the literary devices and its tone,
- understanding situational mood, and
- determining the writer's purpose and drawing inferences about the writer.

*B Reading Comprehension is an important component in education systems*

In most of the education systems, reading comprehension is an important ability for students to develop. For example, the Organisation for Economic Co-operation and Development (OECD) is a group of 34 member countries. OECD members include most of the developed countries like US, UK, German, Japan, France and Australia. OECD has a Programme called International Student Assessment (PISA), as one of the most important measurement about their students' education. Every three years it tests 15-year-old students from all over the world. Reading ability is considered as one of three most important ability students shall build through their development at schools.

*C Reading comprehension ability at different levels*

Reading comprehension ability of human beings are roughly divided into two levels. The shallow level is about the structure and phonemic recognition, sentence structure and etc. The deep processing level requires semantic processing, encode the meaning of words, their similar words and etc.

The following is an example of Australia National Assessment Program - Literacy and Numeracy (NPLAN) test at year 5 [6].

An article of the year 5 NPLAN Test:

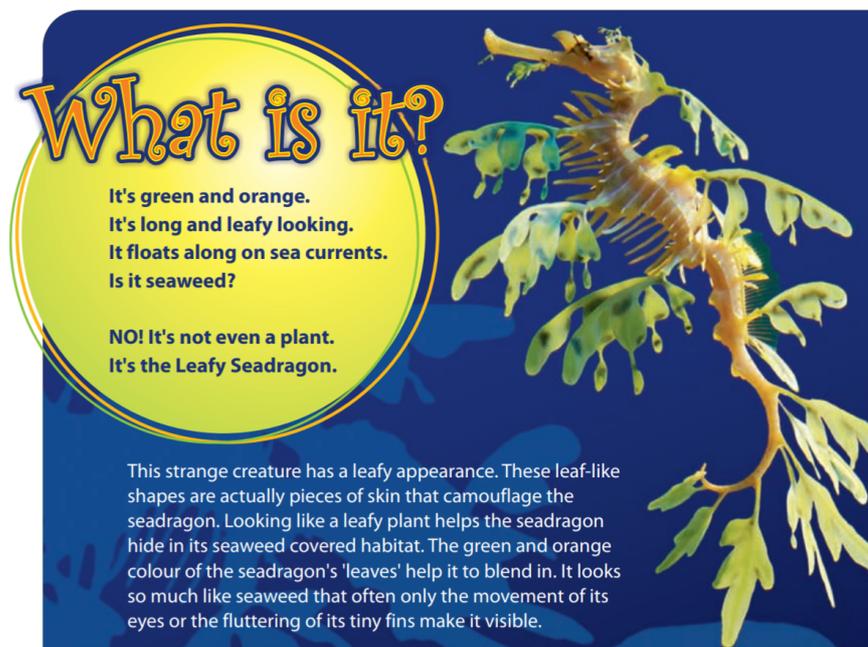

Answer sheet of the year 5 NPLAN test:

**1** A leafy seadragon is
- ◯ a leaf.
- ◯ a fish.
- ◯ a plant.
- ◯ a seahorse.

**2** The section in the yellow circle is written as
- ◯ a riddle.
- ◯ a report.
- ◯ an argument.
- ◯ an explanation.

**3** Which feature makes a seadragon look like seaweed?
- ◯ its fins
- ◯ its skin
- ◯ its eyes
- ◯ its head

As a comparison, the following example is at year 9.  We can see that the requirement on the reading ability increases while the year level progresses.

An article of the year 9 NPLAN Test:

## DROOLING AUTOTROPHS?

Sheldon Cooper, the comically literal-minded character in the TV series *The Big Bang Theory*, would probably respond to the show's witty theme song by pointing out some inaccuracies in the lyrics. For example, the song refers to the beginning of life
5  on Earth as the time when "the autotrophs began to drool". This suggests that an autotroph is a primitive animal when, in fact, plants are the main autotrophs.

*Autotroph* is derived from the Greek *auto* (self) and *trophe* (nutrition). Plants nurture themselves in the sense that they use
10 the energy of the sun to convert carbon dioxide and water into the organic carbon compounds needed for growth and activity. By contrast, *heterotrophs* (animals and fungi) simply consume these pre-produced materials.

Biologists refer to *trophic levels* to analyse ecological systems and
15 how energy flows through their food chains. For the purposes of the theme song of *The Big Bang Theory*, however, the main advantage of the term *autotroph* is that it *sounds* scientific!

Answer Sheet of the Year 9 NAPLAN test:

5  In line 13, the word *these* refers to
   - ○ materials used for production.
   - ○ carbon compounds.
   - ○ animal producers.
   - ○ plant producers.

6  This text could best be used in
   - ○ an advertisement for *The Big Bang Theory*.
   - ○ a review of *The Big Bang Theory*.
   - ○ an encyclopedia for biologists.
   - ○ a popular science magazine.

7  Some unusual organisms belong to **subgroups** of autotrophs and heterotrophs. For example, *chemoautotrophs* are organisms living on chemicals from volcanic vents in the total darkness of the deep sea.

   Choose the name for the subgroup of organisms which
   (i) use power from the sun, BUT
   (ii) get carbon materials for growth from eating other organisms.
   - ○ photoautotroph
   - ○ chemoautotroph
   - ○ photoheterotroph
   - ○ chemoheterotroph

**S**tanford **Q**uestion **A**nswering **D**ataset (SQuAD) is a reading comprehension dataset, consisting of questions posed by crowd workers on a set of Wikipedia articles [3]. Recently, BERT (reference) has made a breakthrough [4]. It led to the beat of human beings at SQuAD, scoring nearly 91 where human users only scored 89+. However, we found that the supreme performance of BERT can be easily broken with specifically designed questions. A number of other researchers had similar findings [5]. It is necessary to develop a model to differentiate different levels of reading comprehension ability.

2. Turing Test

Turing test is a well know test, proposed by Alan Turing in 1950, to determine whether a computer program has gained similar intelligence like human beings [6]. As shown in Fig 2.1, a human questioner will issue a set of questions to both a computer respondent and a human respondent located in two separated rooms. If the human questioner is not able to identify which room is the computer respondent, the computer respondent (algorithm) has passed the Turing Test.

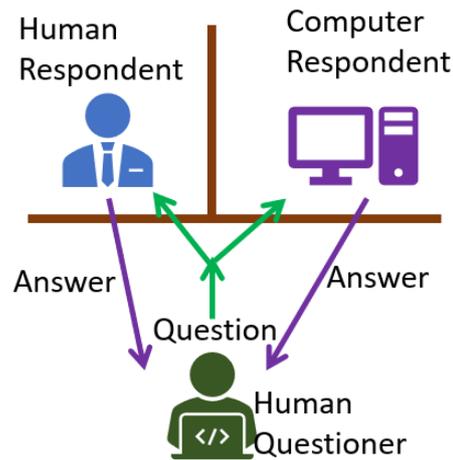

Fig 2.1 Turing Test

For many years questions in Turing test have to be limited for a computer to exhibit human like intelligence. But with the fast development of modern AI technologies, the limitations have been gradually relaxed.

## 3. Reading Comprehension Ability Test (CAT)

Reading Comprehension Ability Test, or reading CAT, is what we proposed to determine whether an AI algorithm has gained similar comprehension ability as human beings. The test setting is similar to Turing Test, but the human questioner is not only issuing a series of questions, but a series of articles and the corresponding question sets, as shown in Fig 3.1.

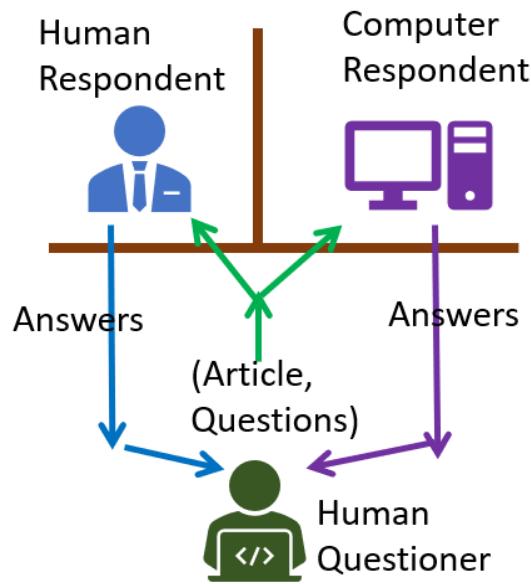

Figure 3.1 CAT - Comprehension Ability Test

A CAT can have $m$ tests. each CAT test, **CAT_test_i,** consists of an article **A$i$** and the corresponding question set **Q$i$**, noted as **CAT_test_i =(A$i$, Q$i$),** where $i$=1, 2, … $m$. The human respondent and the computer respondent need to 'read' the article and answer the corresponding set of questions. If the human questioner is not able to differentiate the computer responder from the human responder, the computer responder has passed the CAT. Its reading comprehension ability has reached the level of the human user.

## 4. CAT Levels

With the recent breakthrough of AI technologies, particularly BERT related algorithms, computers are able to beat human users in the test of the well-known SQuAD dataset. SQuAD covers a wide range of topics. The questions and articles are much less limited as compared to the earlier years' Turing tests. However, after we made some minor modification to the questions, the algorithms failed badly. Even though the modification gave human readers little extra challenges. As earlier discussed, reading comprehension ability are of different levels. The artificial intelligence shall progress level by level until it can beat most of us.

We divided CAT into four levels and will discuss the levels one by one with examples.

*4.1 CAT Level 1 – identifying the facts presented in the article.*

CAT Level 1 requires the ability to identify the facts presented in the article. For example, in the following article extracted from the SQuAD 2.0, there are many facts related to the Swedish school system. The comprehension ability at CAT Level 1 should be able to identify the facts such as
- in Swenden, pupils can choose both private school and private school; or
- 10% of pupils enrolled in private school in 2008; or
- "The knowledge School" is the biggest school chain, and etc.

> In Sweden, pupils are free to choose a private school and the private school gets paid the same amount as municipal schools. Over 10% of Swedish pupils were enrolled in private schools in 2008. Sweden is internationally known for this innovative school voucher model that provides Swedish pupils with the opportunity to choose the school they prefer. For instance, the biggest school chain, Kunskapsskolan ("The Knowledge School"), offers 30 schools and a web-based environment, has 700 employees and teaches nearly 10,000 pupils. The Swedish system has been recommended to Barack Obama.

The followings are two example questions in the SQuAD.

- How many people work for Kunskapsskolan schools?
- Ground Truth Answers: 700

- How many Swedish students were enrolled in public school in 2008?
- Ground Truth Answers: <No Answer>

- As of 2008, about what percentage of Swedish students attended private schools?
- Ground Truth Answers: 10%

For such questions, artificial intelligence is already able to score about 91, which outperformed human users' 89+. By examining the answers, a human questioner is not able to tell whether the comprehension was performed by a human user or an AI algorithm. Therefore, AI has passed CAT Level 1 on SQuAD.

However, if we modify the question by replacing some words with similar ones or its properties referred in the context, BERT algorithms can fail badly. The followings are two examples of such questions:

- As of 2008, about what percentage of primary school students and secondary school students studied in private schools?
- Comprehended Answers: 10% (BERT Prediction: <No Answer>)

- In Sweden, how many pupils were taught in chain schools including school and web environment?
- Comprehended Answers: <No Answer> (BERT Prediction: 10,000)

To be able to answer the corresponding questions, the comprehension ability needs to reach the next level.

*4.2 CAT Level 2 – identifying the facts presented in the article by applying common knowledge of equivalent terms in the context.*

CAT Level 2 requires the ability to identify the facts presented in the article by applying common knowledge on the vocabulary used. Common knowledge on vocabulary includes some basic 'equivalent' relationships of words. For example, we know 'study in school' means 'enroll in school'; or 'pupils' mean 'primary school students and secondary school students':

(Study, context: school) equals (enroll, context: school)

Common knowledge includes the well-known facts and relationships. They are well established and will remain stable in a relatively long period of time. Common knowledge can include some well-known facts but does not include specific information. It should not rely on "new information" to answer the question but the comprehension ability. However, as BERT related algorithms did not model common knowledge, it cannot handle the corresponding questions.

AT CAT Level 2, the common knowledge can be more than synonym. When we describe a property of a substance, in that context, the property and the substance is exchangeable. For example,

The color of snow is white.   Or (snow) equals (white, property: color)

It means snow's color property is white. When the context is color, then snow and white are exchangeable. We can say snow is white, or something's color is snow (white).

If an article includes a sentence that 'a mountain top is full of snow'. By knowing that snow's color property is white, if the question is 'What is the color of the mountain top?', the answer will be 'white'. It is an extension of the common knowledge in the form of vocabulary synonym.

Words in natural language can have many meanings. The meaning often is dependent to the context. Although BERT has little ability to comprehend without common knowledge, its mechanism can be extended to establish context of given common knowledge. Therefore, CAT Level 2 can be well resolved, which we will report in a different article.

*4.3 CAT Level 3 – perform inference with the knowledge expressed in the article or together with the related common knowledge.*

CAT Level 3 requires the ability to perform inference with the knowledge expressed in the article, or with related common knowledge. Human beings can perform inference and draw conclusion accordingly. This is an important part of our reading comprehension ability. Two typical inferences are logical inference and numerical inference.

Logical inference is the connection of logical relationships. For example, if
- 'students' fee at municipal schools is $5,000'; and
- 'pupils pay the same amount for private schools and municipal schools'.

Then we can infer and conclude that
- 'students' fee at private school is $5,000'.

This is a logical inference. Similarly, if
- 'the mountain top is full of snow'; and
- 'snow is cold';

then we can tell that
- 'the mountain top is cold'.

If we can infer through the natural sentences, then the common knowledge can be expressed in natural language, which is much more convenient for its collection than requiring a formal model.

Numerical inference is also very common in our reading comprehension. For example, if
- 'today is Monday' and
- 'the meeting will be three days later',

Then
- 'the meeting will be on Thursday'.

Similarly, if
- since 2008 the Sweden has both private and municipal schools,

then
- the private schools have existed in Sweden for at least 11 years, given this year is 2019.

Although more complex mathematical problem solving requires specialized modelling, general numerical inference is needed in our daily reading comprehension, which appears in schools' literacy tests rather than numeracy tests.

CAT Level 3 requires the artificial intelligence to establish a structure of the sentences and paragraphs to perform inference. Simple classification or encoding will not likely be able to achieve this level of comprehension ability.

*4.4 CAT Level 4 – identifying sentiment (mood, tone, genre), intent and main thought (idea).*

CAT Level 4 requires the ability to identify the sentiment, intent and the main thought of the article. A writer described the weather may be a description of a fact. It can also be a description of the mood of characters. Although there are already interesting research progresses on sentiment analysis. But it is still clearly weaker than human beings' comprehension ability. The following is an example paragraph extracted from a dairy (reading comprehension test).

> I (Ben) woke up this morning, the sun was shining through the curtains and I could smell breakfast cooking downstairs. I jumped out of my bed and threw on my clothes, skipped down to the kitchen.

What mood is the paragraph expressed? Was Ben excited, hungry or in a hurry? Why did the writer describe how Ben get up the bed and went downstairs? We human beings can read, comprehend and answer these questions. To pass CAT Level 4, algorithms will need to achieve the similar ability. The corresponding answers often require the context of the paragraph. From the dairy containing the above paragraph, we know that Ben described how amazing his day was, right from when he got up from bed.

Sentiment analysis techniques are able to perform classification with a large amount of labeled corpus. However, when the sentiment and intent are related to the connection to facts and requires inference, the classification will failure like what BERT did at level 2 and above.

## 5. Conclusion

In this paper we proposed reading Comprehension Ability Test, or CAT. CAT is similar to Turing test but only test computers reading ability. CAT contains four levels of reading comprehension from simple to difficult. We analysis the disadvantage and advantage of the existing computer reading comprehension model (BERT) in four levels. BERT had a great performance at the first level, but did badly at further levels. CAT helps to direct the future work of machine comprehension.